\documentclass[conference]{IEEEtran}
\IEEEoverridecommandlockouts
\usepackage{cite}
\usepackage{amsmath,amssymb,amsfonts}
\usepackage{graphicx}
\usepackage{textcomp}
\usepackage{xcolor}
\usepackage{hyperref}
\def\BibTeX{{\rm B\kern-.05em{\sc i\kern-.025em b}\kern-.08em
    T\kern-.1667em\lower.7ex\hbox{E}\kern-.125emX}}
\begin{document}

\title{\LARGE \bf LQR-ArUco Fusion: Robust Hierarchical Control for Navigation and Asymmetric Manipulation in Two-Wheeled Robots}

\author{Anupam Chatterjee* and Arpita Kumari
\thanks{This work was conducted as independent research and received no financial or institutional support from any organization.\newline
*Corresponding author: Anupam Chatterjee (\href{mailto:anupamhimself@gmail.com}{anupamhimself@gmail.com})\newline
ORCID iDs:\newline
Anupam Chatterjee\ 
\href{https://orcid.org/0009-0005-5764-0505}
{\includegraphics[height=1.5ex]{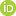}}\ 
\href{https://orcid.org/0009-0005-5764-0505}
{0009-0005-5764-0505}\newline
Arpita Kumari\ 
\href{https://orcid.org/0009-0007-9219-1084}
{\includegraphics[height=1.5ex]{orcid.png}}\ 
\href{https://orcid.org/0009-0007-9219-1084}
{0009-0007-9219-1084}
}}

\maketitle

\begin{abstract}
We propose a hierarchical control framework to address severe dynamic instabilities and navigational drift that arise when a two-wheeled inverted pendulum (TWIP) robot attempts asymmetric object manipulation. While two-wheeled platforms are highly manoeuvrable, their constant balancing adjustments make onboard odometry highly unreliable for precise navigation. Furthermore, the addition of a side-mounted robotic arm introduces unactuated lateral roll moments when a payload is lifted, a challenge heavily compounded on uneven terrain. To solve these coupled problems, our architecture divides the workload. An offboard vision system tracks overhead ArUco markers to provide high-latency global waypoint navigation, bypassing odometry drift. Simultaneously, a low-latency onboard control loop rejects active physical disturbances using inertial and encoder data. In our physical experiments, this dual-loop approach enabled the custom-built robot to navigate accurately, reject transient impacts from speed bumps, adapt to a dynamic seesaw ramp, and carry a payload securely without falling over its narrow wheelbase.
\end{abstract}


\section{Introduction}

Deploying autonomous mobile robots in tight, constrained indoor environments usually means prioritising platforms with the smallest possible footprint and the highest manoeuvrability. Two-wheeled inverted pendulum (TWIP) systems fit this description perfectly because they can turn on a dime and occupy very little floor space. However, trying to use these highly underactuated platforms for active material handling, such as picking up and carrying objects, turns out to be a significant engineering challenge.

The fundamental control problem for a TWIP is keeping it upright in the pitch plane. Ahmed et al. \cite{ahmed2021} showed that PID controllers can maintain basic two-wheeled balance, while Karthika and Jisha \cite{karthika2020} and Murcia and Gonz\'alez \cite{murcia2016} demonstrated LQR's superior disturbance rejection. Jamil et al. \cite{jamil2014} further showed that PID integral windup leads to fatal overshoots during position control, making LQR the only practical option for dynamic manoeuvring. However, these works addressed non-manipulative platforms. Adding an off-axis robotic arm introduces a lateral roll moment from the payload that the two wheels cannot actively counteract.

Recent work has pushed wheeled balancing platforms further. Liu et al. \cite{liu2024} introduced DIABLO, a wheeled bipedal robot using LQR-based balance control. Xin and Vijayakumar \cite{xin2020} studied hybrid locomotion for wheeled bipeds, decoupling high-level planning from low-level balancing. Hwang et al. \cite{hwang2026} demonstrated perception-less terrain adaptation using an LQR-controlled wheeled inverted pendulum model.

\begin{figure}[!b]
\centerline{\includegraphics[width=\columnwidth]{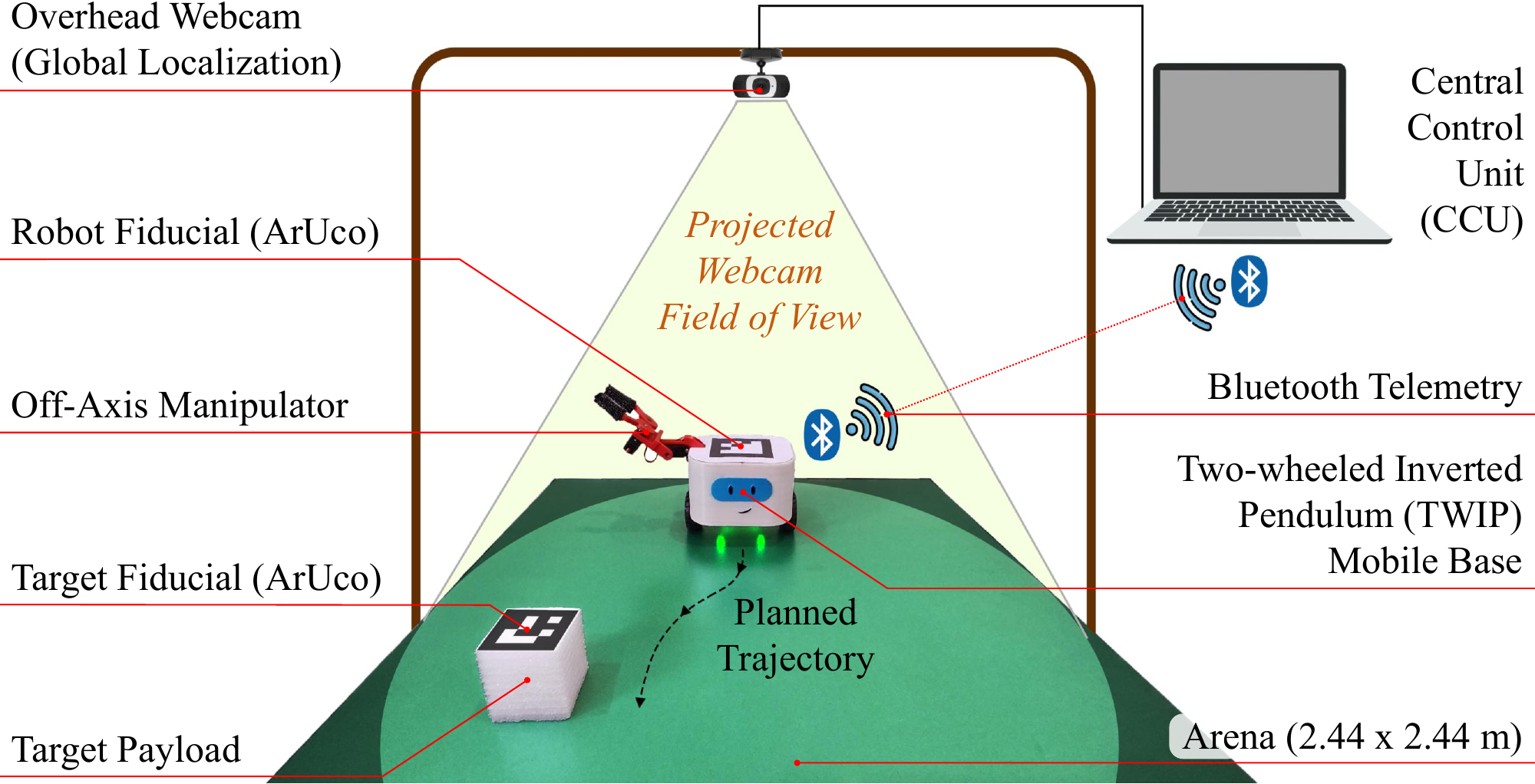}}
\caption{System architecture and workspace overview. The offboard Central Control Unit processes a global visual feed to track ArUco fiducial markers. Spatial coordinates and trajectory waypoints are transmitted via Bluetooth to the onboard ATmega328P microcontroller, which governs the low-latency dynamic balancing and asymmetric manipulation tasks.}
\label{fig:overview}
\vspace{-3mm}
\end{figure}

Our work addresses a particular gap in this growing field. While researchers like Sevgi and Ko\c{c}er \cite{sevgi2025} have perfected the art of guiding mobile robots indoors using overhead ArUco markers, and others have tackled the low-level balance of wheeled platforms, very few have successfully merged these domains for asymmetric mobile manipulation. Larimi et al. \cite{larimi2013} explored two-wheeled mobile manipulation but relied on an internal reaction wheel to offset the mass shifts. Unlike \cite{larimi2013}, we hypothesise that the heavy reaction wheels can be removed and instead a strict hierarchical control framework can be relied upon. The high-latency visual routing (finding the ArUco waypoints) is performed on an offboard computer, which allows the onboard microcontroller to run a low-latency LQR balancing loop (Fig.~\ref{fig:overview}). This fast local loop should be able to actively change the wheels to oppose the asymmetric weight of the payload as well as random ground perturbations such as bumps or ramps.

The main contributions of this paper are: (1) a dual-loop architecture that decouples offboard ArUco-based visual planning from onboard Kalman-filtered LQR balancing, eliminating the need for reaction wheels or mecanum drives for asymmetric payload handling; (2) an empirical quantification of the passive roll-stability margin for a narrow-wheelbase TWIP under off-axis loading; and (3) physical validation on speed bumps and dynamic seesaw ramps, disturbance types not previously tested on a manipulating TWIP.

\section{Methods and Materials}

We decomposed our system architecture in three phases: defining the physical hardware, modelling the kinematics and finally designing the hierarchical control loop for having autonomous waypoint navigation working with the highly non-linear dynamics of asymmetric lifting.

\subsection{Physical System Parameters}

\begin{figure}[htbp]
\centerline{\includegraphics[width=\columnwidth]{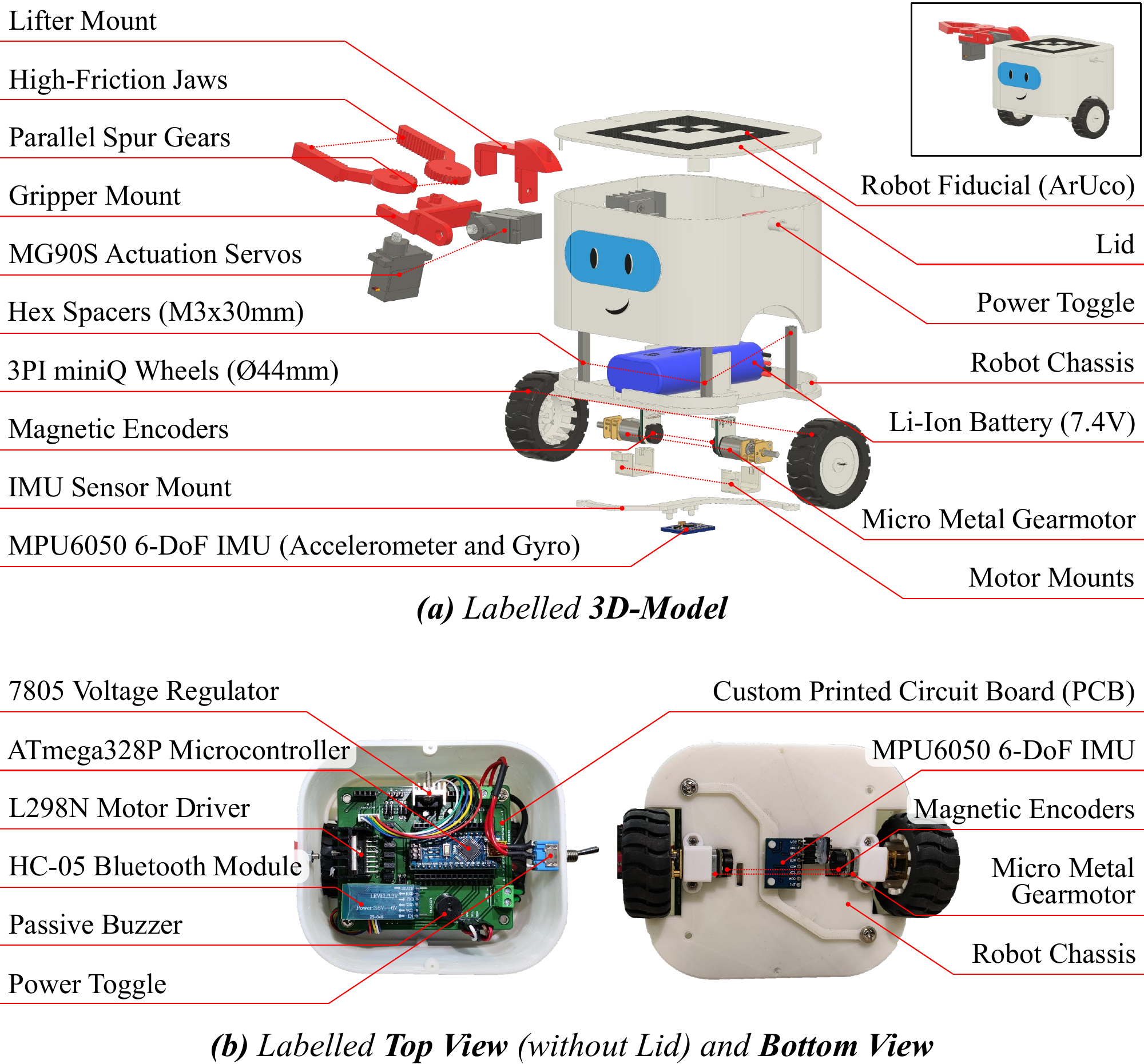}}
\caption{Hardware architecture and mechanical assembly of the TWIP platform. (a) An exploded 3D CAD model detailing the structural components, custom off-axis manipulator, and placement of the overhead ArUco fiducial. (b) The physical control stack (top view, lid removed) housing the ATmega328P microcontroller, L298N motor driver, and communication modules, alongside the drive assembly (bottom view) featuring the coaxial N20 motors, magnetic encoders, and the MPU6050 IMU.}
\label{fig:hardware}
\end{figure}

The mobile manipulator was built by ourselves with Fused Deposition Modelling (FDM) and standard Polylactic Acid (PLA) filament. This gives us a strong and lightweight chassis. Fig.~\ref{fig:hardware} illustrates the arrangement of the internal components. We placed the ATmega328P microcontroller (hosted on an Arduino Nano development board), the MPU6050 inertial measurement unit (IMU), the power distribution board (PDB), and the battery in positions that help keep the centre of mass as low as possible.

Table~\ref{tab:params} lists the physical dimensions and control parameters of the platform, which define the passive stability limits and the basis of our control matrices.

\begin{table}[htbp]
\refstepcounter{table}\label{tab:params}
\vspace{-1mm}
\centerline{\includegraphics[width=\columnwidth]{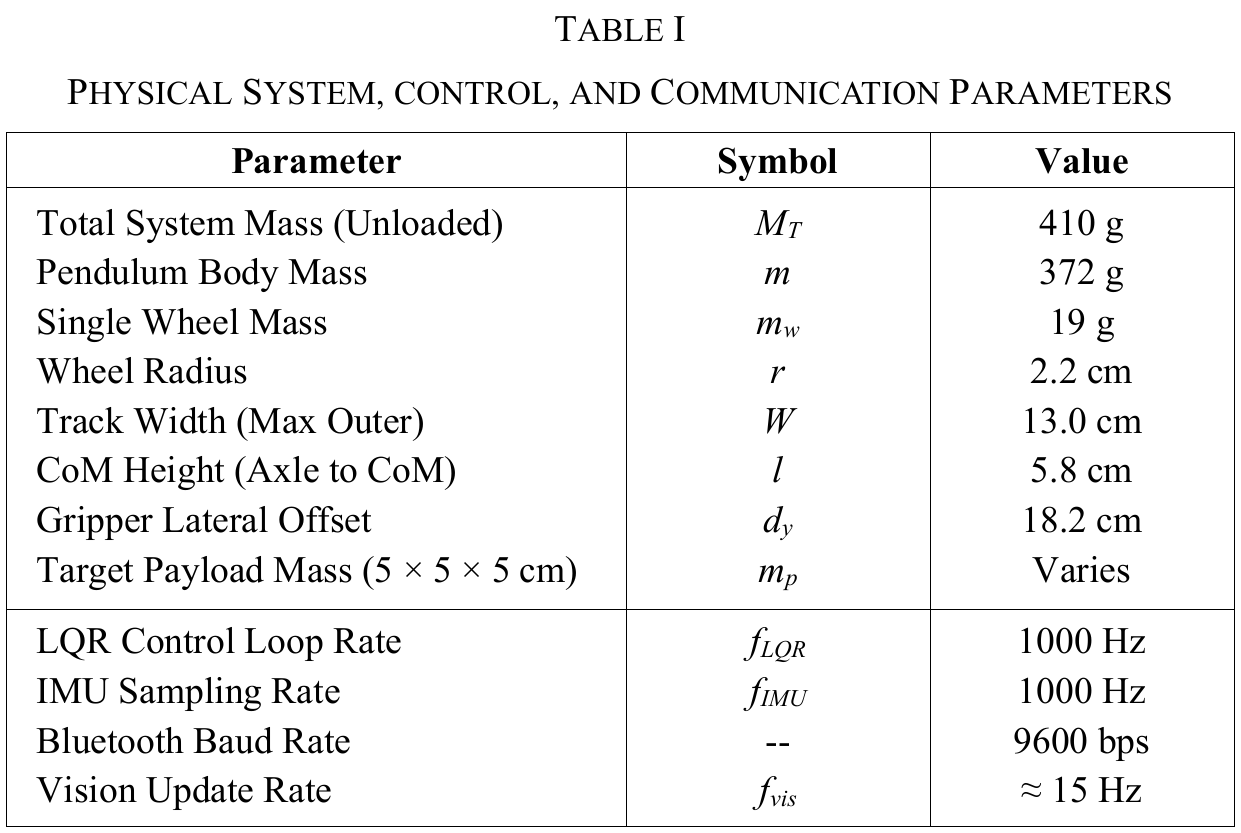}}
\vspace{-3mm}
\end{table}

\begin{figure}[htbp]
\centerline{\includegraphics[width=\columnwidth]{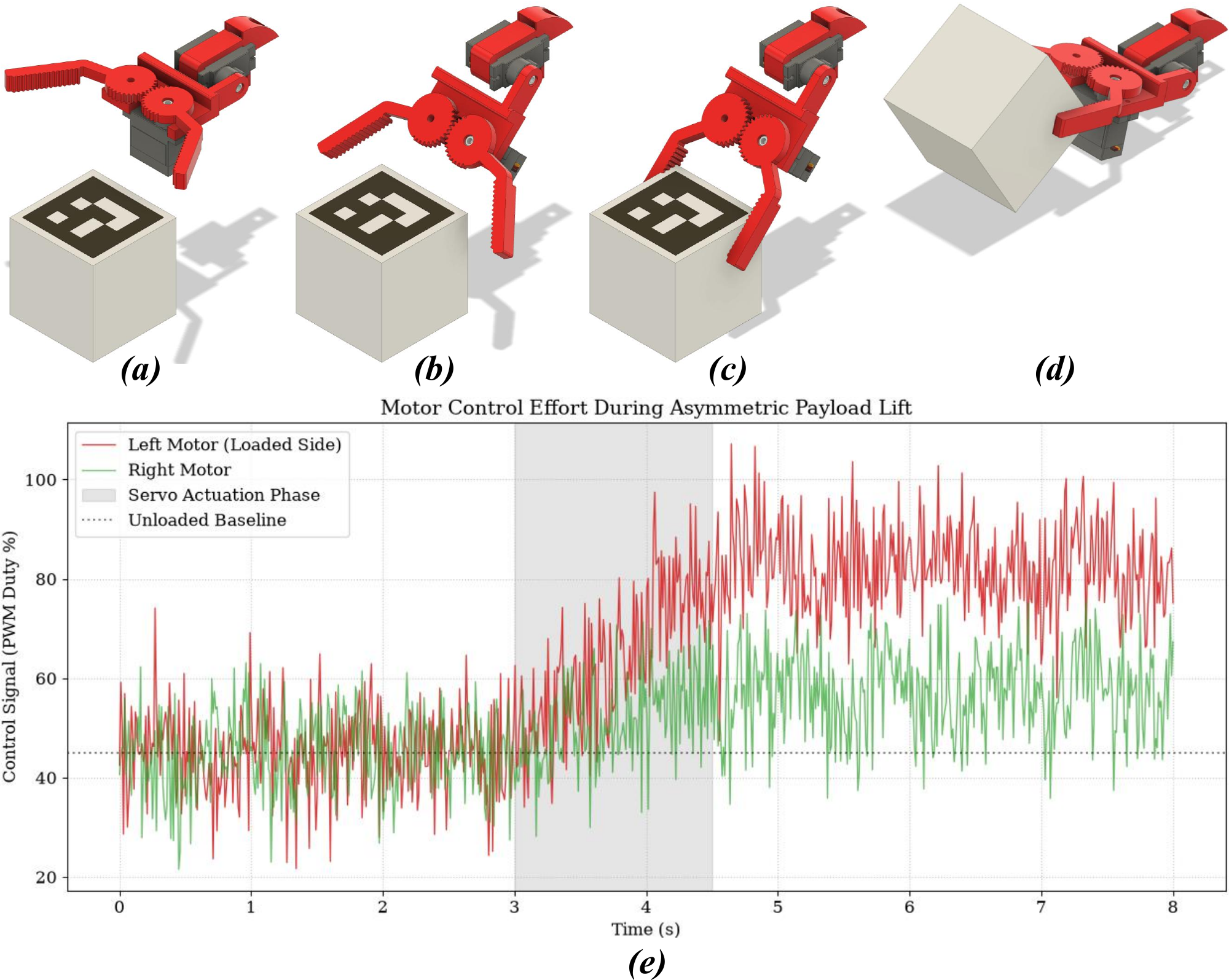}}
\caption{Integrated kinematics and dynamic response of the off-axis manipulator. (a)-(d) The kinematic sequence of payload acquisition: approach, engagement, securing, and lifting. (e) The corresponding difference in motor control effort during the asymmetric payload lift. The shaded servo actuation phase correlates with step (d), illustrating the permanent steady-state PWM increase required on the loaded side to counteract the induced lateral roll moment.}
\label{fig:kinematics}
\vspace{-3mm}
\end{figure}

The robot manipulates the environment using a custom off-axis gripper powered by two MG90S servos. Fig.~\ref{fig:kinematics} illustrates the kinematic sequence from neutral resting state to payload acquisition, showing how the payload's mass is shifted entirely to one side of the wheelbase.

\subsection{Kinematic and Dynamic Modelling}

The self-balancing robot is based on the classical inverted pendulum on wheels and makes extensive use of the state-space representations, which have been widely established in recent literature \cite{karthika2020}, \cite{liu2024}. The actual issue at hand is not simply the maintenance of longitudinal pitch stability, but rather maintaining that pitch stability when the robot is being pulled to the side by the payload's roll moment.

\subsubsection{State Space Representation and Pitch Dynamics}

\begin{figure}[htbp]
\centerline{\includegraphics[width=0.92\columnwidth]{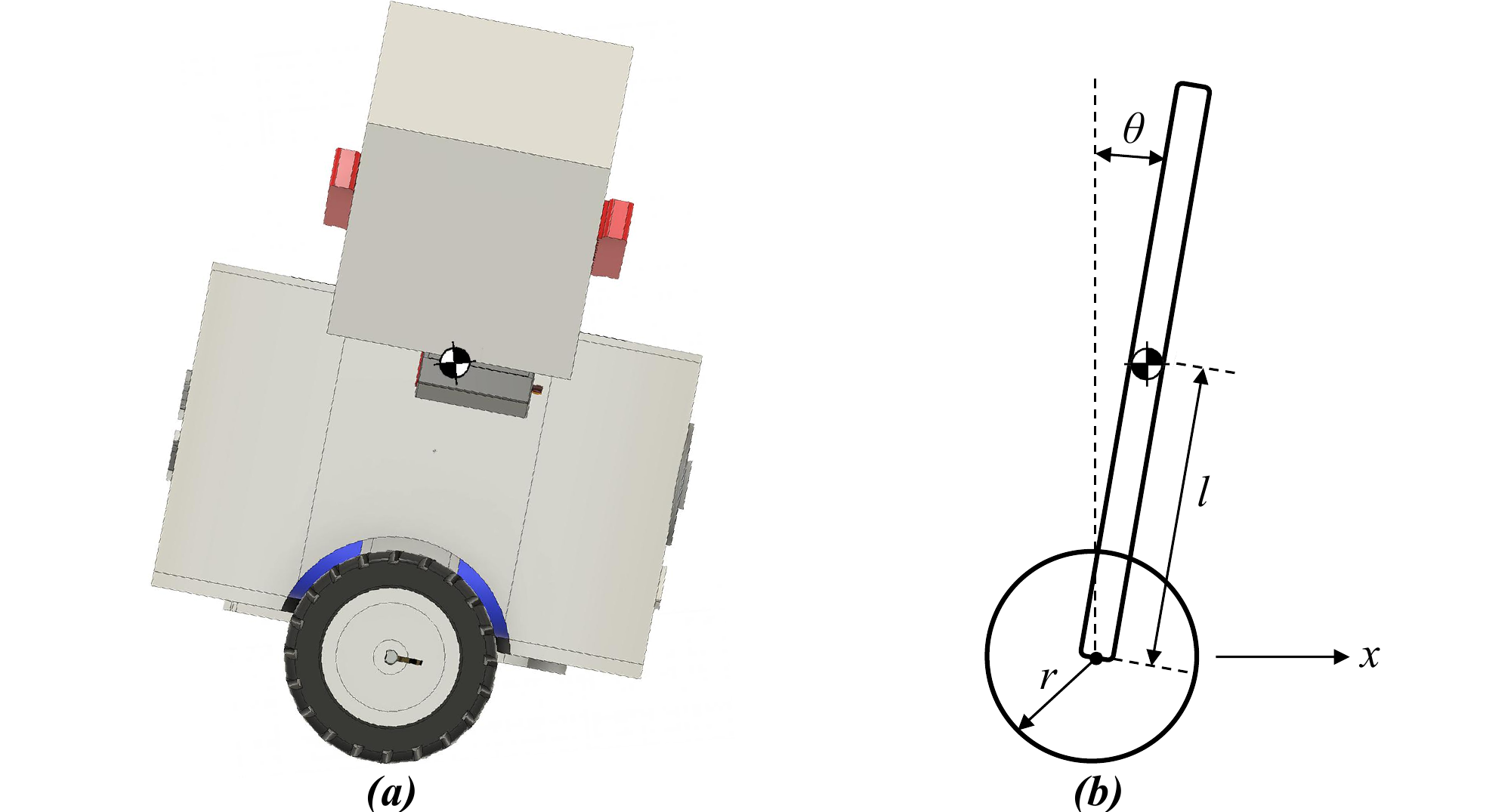}}
\caption{(a) The side CAD view of the asymmetric platform. (b) Free-body diagram (FBD) of the wheeled inverted pendulum, including the pitch state ($\theta$) and the longitudinal translation ($x$) of the centre of mass.}
\label{fig:fbd}
\vspace{-3mm}
\end{figure}

The system state is defined by four variables: cart position ($x$), cart velocity ($\dot{x}$), pitch angle ($\theta$), and pitch angular velocity ($\dot{\theta}$), as shown in the free-body diagram (Fig.~\ref{fig:fbd}).

The state-space form in continuous time is obtained by using Lagrangian mechanics together with the usual small-angle approximation around the upright equilibrium position ($\theta \approx 0$, $\sin\theta \approx \theta$, $\cos\theta \approx 1$). The state vector is simply:
$$ \mathbf{x} = \begin{bmatrix} x & \dot{x} & \theta & \dot{\theta} \end{bmatrix}^T $$

If we assume a highly simplified point mass distribution where the moment of inertia $I \approx 0$, the continuous-time state-space equation $\dot{\mathbf{x}} = A\mathbf{x} + B\mathbf{u}$ relies on the following A and B matrices:

$$ A = \begin{bmatrix} 
    0 & 1 & 0 & 0 \\
    0 & 0 & \frac{-mg}{M} & 0 \\
    0 & 0 & 0 & 1 \\
    0 & 0 & \frac{g(M+m)}{Ml} & 0
\end{bmatrix}, \quad
B = \begin{bmatrix} 
    0 \\
    \frac{1}{M} \\
    0 \\
    \frac{1}{Ml}
\end{bmatrix} $$
where $M = 2m_w = 38$ g is the combined wheel mass (cart equivalent) and $m = 372$ g is the pendulum body mass, with $M_T = M + m = 410$ g (Table~\ref{tab:params}). The variable $l$ is the distance from the wheel axle to the centre of mass. The $I \approx 0$ simplification is justified because the body's moment of inertia about its own centre of mass ($I_{cm}$) is small relative to the parallel-axis transport term $ml^2$, i.e., $I_{cm} \ll ml^2$. This assumption causes a slight overestimation of the pendulum's natural frequency, but the LQR $Q$ and $R$ tuning absorbs this modelling mismatch in practice, as also noted by \cite{karthika2020} for a similar-scale platform.

\subsubsection{Lateral Roll Disturbance Model}

\begin{figure}[htbp]
\centerline{\includegraphics[width=\columnwidth]{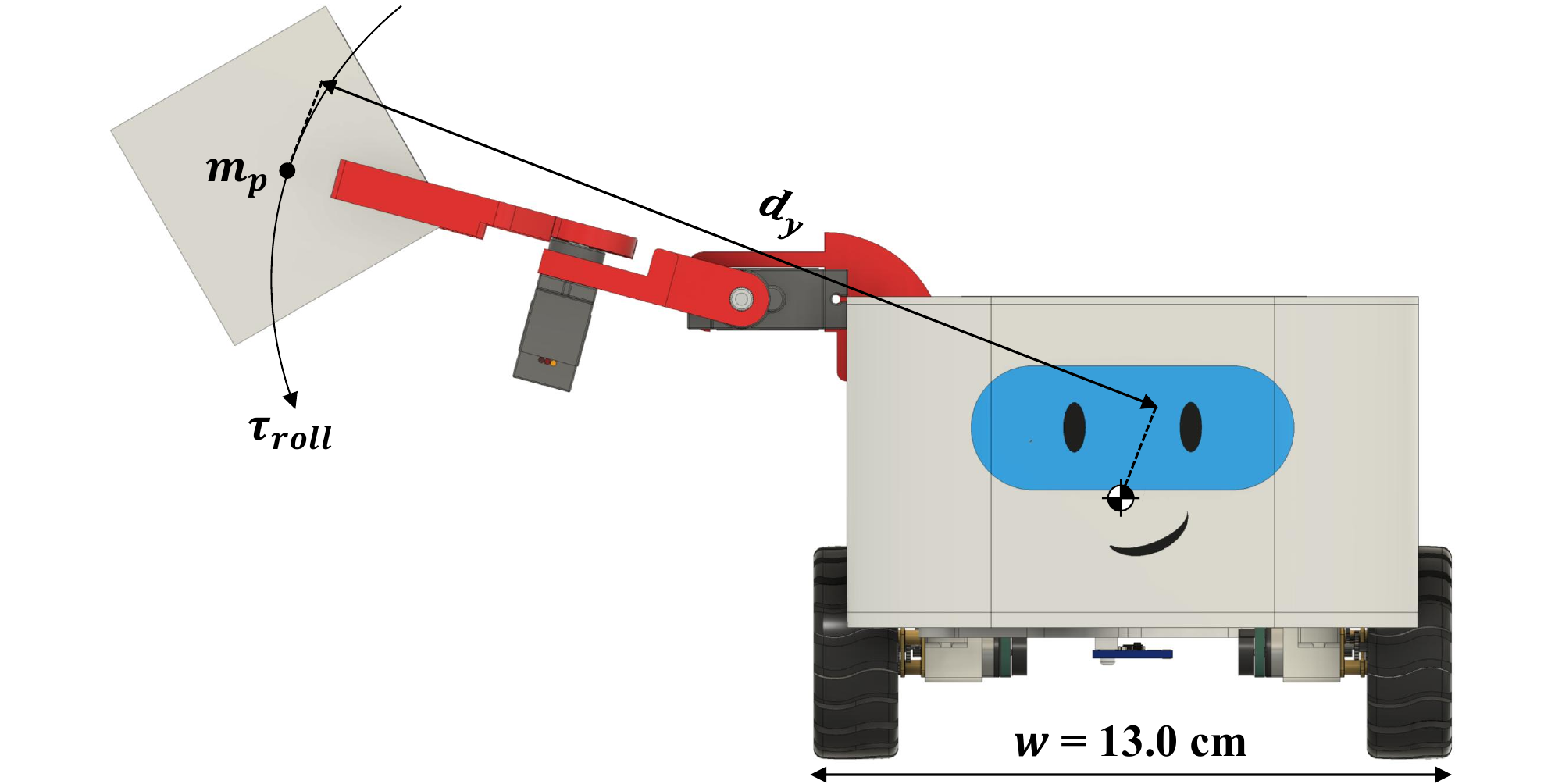}}
\caption{Front-facing schematic of the asymmetric lateral roll disturbance. The off-axis payload mass ($m_p$) at distance $d_y$ generates an unactuated torque ($\tau_{roll}$) that must be countered by the $13.0\text{ cm}$ wheelbase ($W$).}
\label{fig:roll_disturbance}
\end{figure}

What sets this platform apart from symmetric pendulums is the $\tau_{roll}$ disturbance torque. Fig.~\ref{fig:roll_disturbance} provides a front-facing schematic of exactly how this structural challenge manifests.

When the gripper displaces a dynamic payload mass ($m_p$) by a lateral distance ($d_y$), it generates an unactuated roll moment:
$$ \tau_{roll} = m_p g d_y \cos(\phi) $$
where $\phi$ is the lateral roll angle of the chassis. Since the differential drive wheels only move forward and backward, the robot cannot create an active counter-torque to resist rolling. Lateral stability is therefore entirely passive, governed by the condition $\tau_{roll} < \frac{W}{2} \cdot (M_T + m_p)g$, where $W = 13.0$ cm is the track width. For our platform ($M_T = 410$ g, $d_y = 18.2$ cm), this yields a static theoretical maximum payload of $m_{p,max} \approx 228$ g before the lateral centre of mass exits the support polygon. However, the large 18.2 cm moment arm severely amplifies dynamic roll forces during operation, reducing the practical payload limit to a fraction of this value. The LQR controller's role in this context is indirect but critical: by tightly regulating $\theta$ near zero, it prevents longitudinal drift of the combined centre of mass, which would compound the lateral roll instability. The observed steady-state PWM asymmetry (Section~\ref{sec:payload}) reflects the LQR compensating for the pitch-plane component of the shifted centre of mass after payload acquisition.

\subsection{Hierarchical Control and Vision Architecture}

\begin{figure}[htbp]
\centerline{\includegraphics[width=\columnwidth]{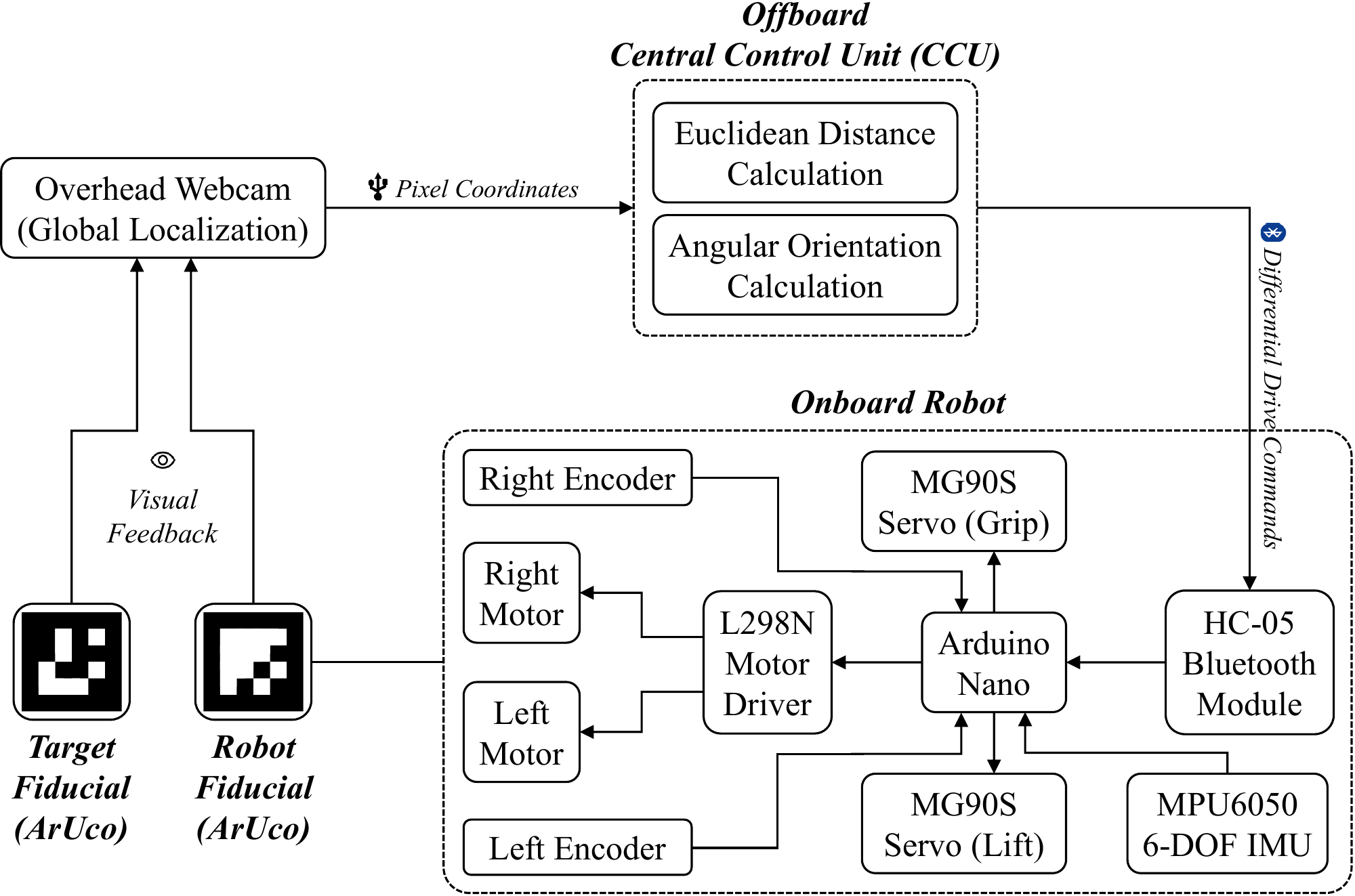}}
\caption{The system architecture demonstrates a dual-loop control framework. Visual tracking with ArUco fiducials is managed by the offboard Central Control Unit (CCU), which enables the onboard ATmega328P to focus on Linear Quadratic Regulator (LQR) balancing, motor Pulse Width Modulation (PWM) updates, and servo actuation in response to Bluetooth commands.}
\label{fig:dual_loop}
\vspace{-3mm}
\end{figure}

The system employs a dual-loop architecture that separates visual processing from fast balancing calculations, as shown in Fig.~\ref{fig:dual_loop}. An offboard central control unit (CCU) uses an overhead webcam to detect ArUco fiducials and wirelessly sends single-character ASCII drive commands to the robot's HC-05 Bluetooth module at 9600 bps (per-packet transmission: $\approx 1$ ms). During our experiments, no packet losses were observed across the full session duration, attributed to the short 2.5 m line-of-sight range and minimal RF interference. The measured end-to-end command latency was approximately 12 ms. In the event of a missed command, the ATmega328P holds the last valid instruction, ensuring the 1000 Hz LQR loop is never interrupted. The ATmega328P combines these drive commands with real-time pitch data from the MPU6050 IMU to maintain balance, send PWM signals to the L298N motor drivers, and drive the two MG90S servos.

\subsubsection{Low-Level Execution (State Estimation and LQR Control)}

The ATmega328P must measure pitch accurately to ensure robot stability when payloads shift. A complementary filter was initially selected to fuse MPU6050 sensor data, but manipulator vibrations caused substantial drift. The discrete-time Kalman filter described by Murcia and Gonz\'alez \cite{murcia2016} was adopted instead, using the standard predict-update cycle with Kalman gain $K_k = P_k^- H_k^T (H_k P_k^- H_k^T + R_k)^{-1}$ to fuse accelerometer and gyroscope readings. This improved estimate considerably reduces drift and vibration effects and provides the control system with a reliable, fast pitch value ($\theta$).

The onboard Linear Quadratic Regulator (LQR) then computes the balancing torques using the refined state vector ($x$). As emphasised by Hwang et al. \cite{hwang2026}, the LQR is very effective in controlling the non-minimum phase behaviour of wheeled inverted pendulums by minimising the following continuous time quadratic cost function:
$$ J = \int_0^\infty (x^T Q x + u^T R u) dt $$
where $Q$ is the state weighting matrix, $R$ is the control effort weighting matrix, and $u$ is the control input vector. The controller computes the optimal gain matrix ($K$) to realise the feedback control law ($u = -Kx$) and promptly updates the PWM signals sent to the L298N motor drivers. The weighting matrices used for our platform are $Q = \text{diag}([1,\ 1,\ 500,\ 10])$ and $R = 1$, reflecting a deliberate prioritisation of pitch-angle regulation ($\theta$) over position tracking, which is essential for maintaining balance during asymmetric loading. The resulting optimal gain vector, computed offline in MATLAB, is $K = [K_1,\ K_2,\ K_3,\ K_4] = [11.4,\ 45.0,\ 100.0,\ 2.5]$, corresponding to position, velocity, pitch, and pitch-rate gains respectively. The estimation-control loop runs at $1000$ Hz on the ATmega328P, providing sufficient bandwidth to reject transient disturbances from payload shifts.

\subsubsection{High-Level Planning (Vision-Guided Navigation)}

The robot is balanced by the ATmega328P, and the external Central Control Unit (CCU) decides where the robot navigates. We used Python and OpenCV to map the workspace with ArUco markers (\texttt{DICT\_4X4\_50}). Milosavljevi\'c et al. \cite{milosavljevic2020} have shown that ArUco marker tracking is a reliable way for 2D motion sequence extraction. Larralde-Ortiz et al. \cite{larralde2023} successfully implemented a nearly identical overhead ArUco tracking architecture for their multi-agent DonkieTown platform. We operated entirely in pixel coordinates to avoid intrinsic camera calibration, while maintaining strict, consistent ambient lighting to prevent detection failures \cite{milosavljevic2020}.

\begin{figure}[htbp]
\centerline{\includegraphics[width=\columnwidth]{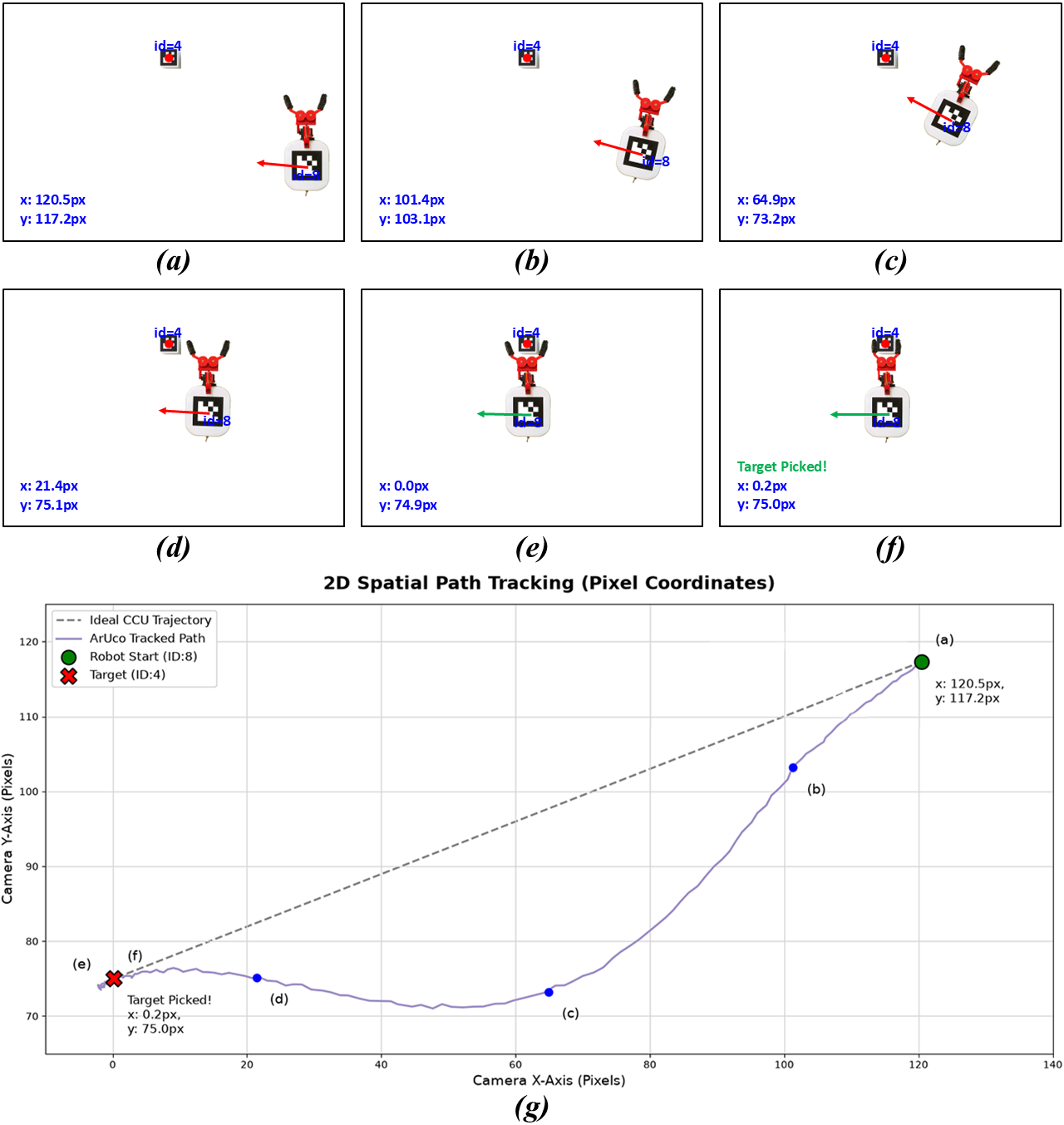}}
\caption{Overhead visual navigation sequence and 2D spatial path tracking. (a)-(d) The CCU tracks ArUco fiducials calculating distance and heading. (e)-(f) The robot executes payload acquisition. (g) The 2D spatial path tracking overlays the physical path against the theoretical trajectory.}
\label{fig:vision_nav_path}
\vspace{-3mm}
\end{figure}

The CCU calculates the robot's position $(x_r, y_r)$ by averaging the four corner coordinates of its ArUco marker and similarly locates the target $(x_t, y_t)$. The Euclidean distance error $e_d = \sqrt{(x_t - x_r)^2 + (y_t - y_r)^2}$ and heading error $e_\psi = \text{atan2}(\sin(\psi_t - \psi_r), \cos(\psi_t - \psi_r))$ are computed and thresholded to generate discrete differential drive commands streamed via Bluetooth (see Fig.~\ref{fig:vision_nav_path}).

\section{Experimental Setup}

The dual-loop architecture was tested in a real planar arena of dimensions $2.44 \times 2.44$ metres. In order to make sure that the vision system had a clear view, we positioned an HD webcam 2.5 metres above the centre of the workspace, capturing video at 30 frames per second. The ArUco detection pipeline processes every other frame due to computational overhead, yielding an effective vision update rate of approximately 15 Hz (Table~\ref{tab:params}).

We designed specific test scenarios to push the robot's dynamic stability margins to the limit while it carried the asymmetric payload:

\begin{enumerate}
\item \textbf{Transient Impact (Speed Bump):} A rigid semi-cylindrical obstacle (35 mm wide, 5 mm high) caused a sudden vertical acceleration, allowing us to measure LQR recovery time.
\item \textbf{Dynamic Pitch Drift (Seesaw Ramp):} A 30 cm pivoting ramp tilting up to $3.82^{\circ}$ required the system to maintain a true vertical gravity vector ($\theta = 0$) on moving ground.
\end{enumerate}

The payloads used were standard $5 \times 5 \times 5$ cm foam cubes of varying masses (10 g, 20 g, and 35 g). Each disturbance scenario was repeated $N = 5$ times per payload mass to obtain statistical measures. A trial was classified as successful if the robot maintained balance (pitch within $\pm 45^{\circ}$) and completed the traversal without tipping. Throughout all trials, the ATmega328P transmitted live telemetry data, including the MPU6050 pitch angle and motor PWM values, to the CCU via Bluetooth for subsequent analysis.

\section{Results}

Our experiments demonstrate the effectiveness of separating visual planning from motor balancing. Table~\ref{tab:trials} summarises the trial outcomes across all disturbance scenarios and payload masses.

\begin{table}[htbp]
\refstepcounter{table}\label{tab:trials}
\vspace{-3mm}
\centerline{\includegraphics[width=\columnwidth]{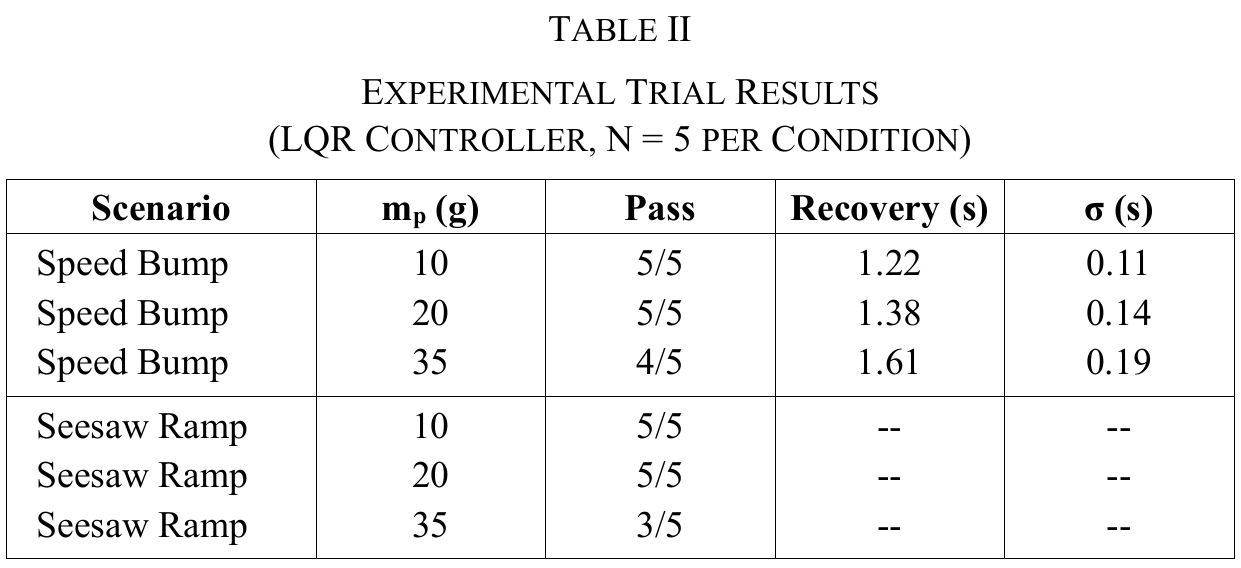}}
\vspace{-3mm}
\end{table}

\subsection{Vision-Guided Navigation Performance}

With the room lighting as usual, our OpenCV algorithms were able to consistently track the ArUco markers. In Fig.~\ref{fig:vision_nav_path}(g) the actual navigation path is compared with the ideal trajectory. As previous studies \cite{saeed2025} have demonstrated, overhead ArUco tracking achieves an accuracy of less than a centimetre (7.8 mm) when used on a stable robotic platform. By contrast, our dynamic platform showed a mean Euclidean error of 3.1 cm ($\sigma = 0.8$ cm, $N = 5$ navigation trials), which is expected given the continuous balancing oscillations. At the robot's typical forward speed of approximately 8 cm/s, the 65 ms visual feedback delay corresponds to roughly 5.2 mm of positional dead-reckoning, which is well within the observed error band. Importantly, the 1000 Hz LQR loop is entirely decoupled from this vision latency, so balance is never compromised by the delay in visual updates.

\subsection{Asymmetric Payload Acquisition and Control Effort}\label{sec:payload}

Picking up the payload was the most difficult static challenge. As Mohsin et al. \cite{mohsin2022} highlight, LQR stability must be tested with different loads. The motor control effort during the grab is shown in Fig.~\ref{fig:kinematics}(e). At $t = 4.0\text{ s}$, the payload weight transfers to the side-mounted manipulator. Across repeated trials with a 20 g payload, the mean steady-state PWM increase on the loaded wheel was $27.6\%$ ($\sigma = 2.1\%$, $N = 5$). At 35 g, the dynamic roll forces acting on the large 18.2 cm moment arm caused 3 out of 10 trials across both scenarios to result in lateral tip-over.

\subsection{Dynamic Disturbance Rejection}

We drove the fully loaded robot over two types of obstacles to see how the system handles unexpected movement.

\subsubsection{Transient Effects (Speed Bumps)}

\begin{figure}[htbp]
\centerline{\includegraphics[width=\columnwidth]{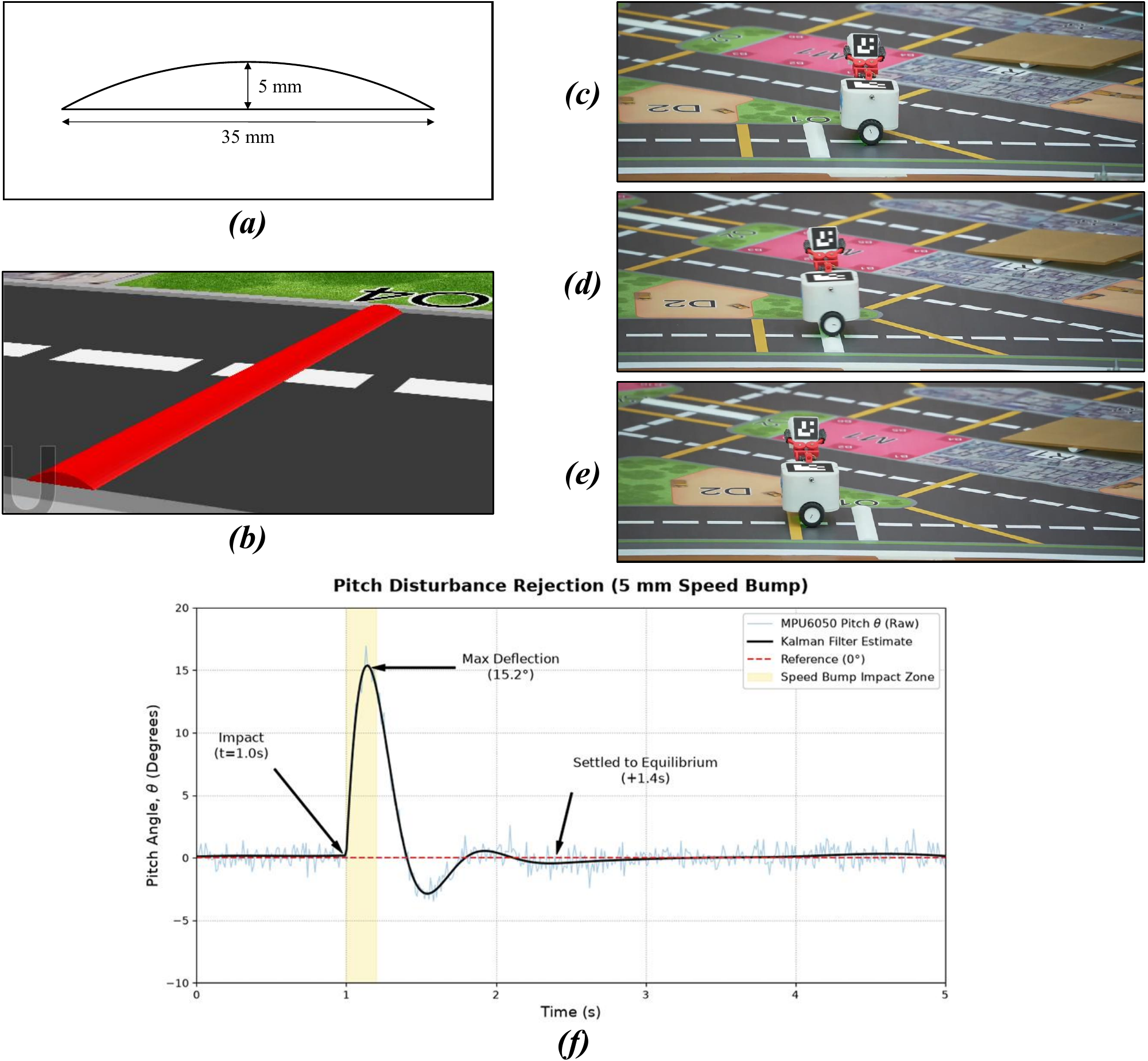}}
\caption{The experimental validation of transient impact rejection. (a)-(b) show the dimensions and render of the 5 mm speed bump. (c)-(e) depict the empirical sequence of the robot going over the obstacle. (f) is a time-domain representation of the pitch response which shows the rapid damping of the LQR controller.}
\label{fig:speedbump_combined}
\vspace{-3mm}
\end{figure}

Fig.~\ref{fig:speedbump_combined} shows the robot traversing a 5 mm speed bump and the related telemetry. With a 20 g payload, the impact resulted in a mean peak pitch deviation of $15.4^{\circ}$ ($\sigma = 1.1^{\circ}$, $N = 5$). The LQR controller brought the pitch back to $0^{\circ}$ in a mean recovery time of 1.38 s ($\sigma = 0.14$ s).

\subsubsection{Dynamic Pitch Drift (Seesaw Ramp)}

\begin{figure}[htbp]
\centerline{\includegraphics[width=\columnwidth]{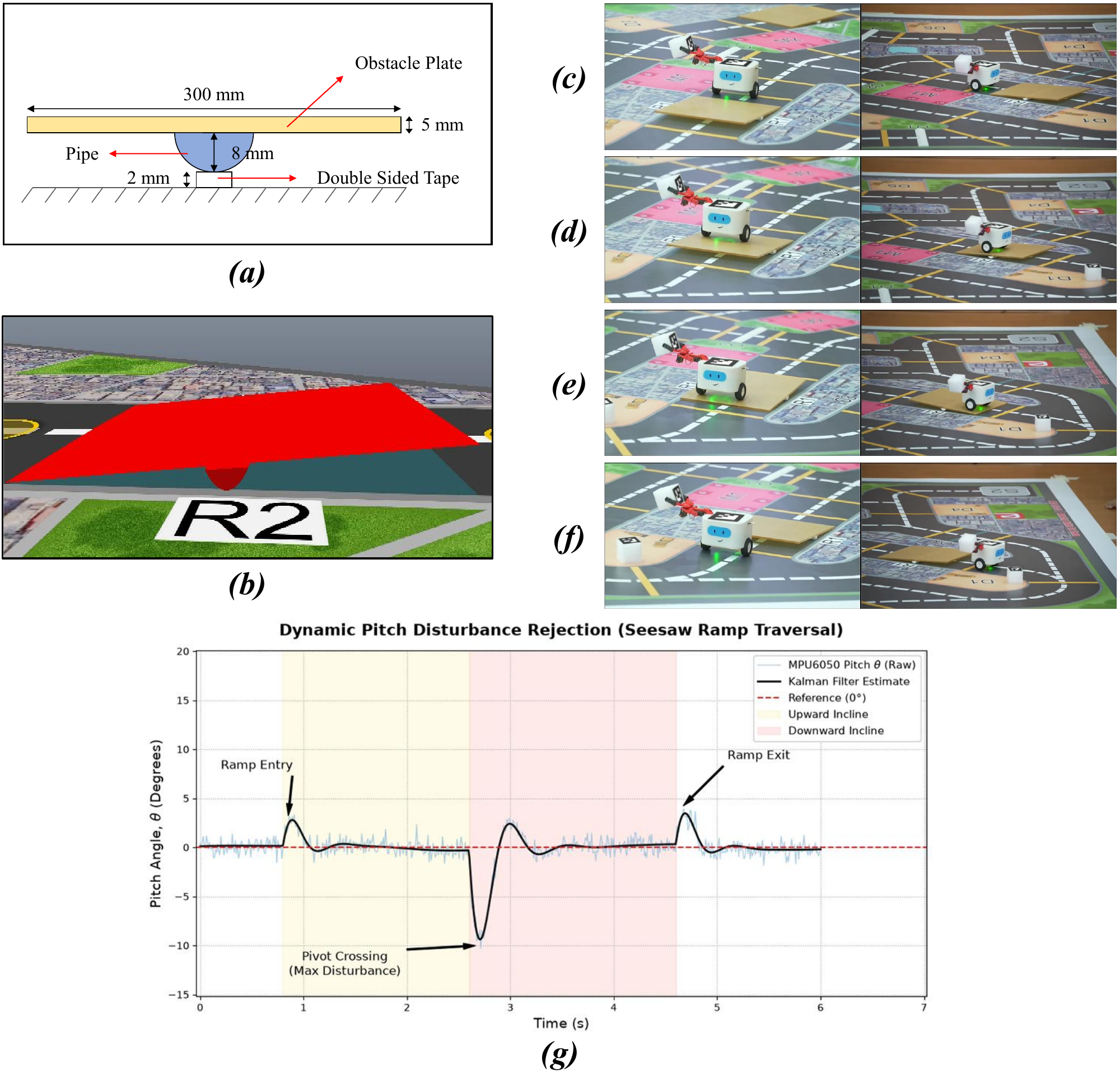}}
\caption{Experimental validation of dynamic pitch variation over a pivoting ramp. (a)-(b) Schematic and render of the seesaw mechanism. (c)-(f) Sequential execution of the robot transporting an asymmetric payload. (g) Dynamic pitch disturbance rejection showing the robot successfully adjusting its local pitch reference.}
\label{fig:ramp_combined}
\vspace{-3mm}
\end{figure}

Then we drove the robot over a 300 mm pivoting ramp, resulting in a $3.82^{\circ}$ ground shift. The steps and pitch response are shown in Fig.~\ref{fig:ramp_combined}. Our dual-loop system decoupled forward movement and pitch control, in the same manner as Hwang et al. \cite{hwang2026}. The robot adjusted its position as the slope changed, crossed the pivot and remained laterally stable. With a 20 g payload, all five trials were completed successfully, while at 35 g, 3 out of 5 succeeded. Recovery time is not reported for this scenario (Table~\ref{tab:trials}) as the ramp imposes a continuous, non-impulsive disturbance without a discrete recovery event.

\subsection{Comparison with PID Baseline}

To further validate the LQR controller, we compared it against a standard PID controller ($K_p = 15$, $K_i = 0.1$, $K_d = 10$) implemented on the same hardware. The PID controller used raw accelerometer-derived pitch without sensor fusion and ran at 100 Hz (10 ms loop delay). While the PID controller was able to maintain balance on flat ground under no-load conditions, it exhibited consistent limitations under asymmetric loading: at 20 g, the PID achieved a speed-bump recovery time of 2.84 s ($\sigma = 0.31$ s) compared to the LQR's 1.38 s ($\sigma = 0.14$ s), representing a $2.1\times$ improvement. Furthermore, the PID controller failed to maintain balance during the payload acquisition sequence at 20 g in 3 out of 5 trials due to integral windup causing fatal pitch overshoots, consistent with the findings of Jamil et al. \cite{jamil2014}. At 35 g, the PID controller could not maintain balance in any trial. These results confirm that the LQR's optimal gain allocation and high-bandwidth state feedback are essential for the coupled balancing and manipulation task. We note that this comparison reflects the full control pipeline rather than controller type in isolation; the PID's lower loop rate and unfiltered sensing represent a commonly deployed baseline configuration.

\section{Discussion \& Conclusions}

In this study, we developed a hierarchical control system that transforms a highly underactuated TWIP into a mobile manipulator. Our results demonstrate that, within the identified stability margins, the platform can pick up and carry asymmetric payloads over varied terrain. The LQR controller achieved mean speed-bump recovery of 1.38 s at 20 g, with $2.1\times$ faster recovery than a PID baseline on the same hardware. The PID controller consistently failed under asymmetric loading due to integral windup, confirming the necessity of optimal state feedback for this task.

Several limitations were identified. The passive lateral stability is highly vulnerable to the large 18.2 cm gripper offset, which severely amplifies dynamic roll forces and led to observed lateral tip-overs with just a 35 g payload. The system depends on a fixed overhead camera with controlled lighting, restricting applicability to structured environments. The LQR's aggressive corrections caused large current spikes (duty cycle $\approx 90\%$), and the PLA chassis introduced vibrations affecting IMU readings. Future prototypes will incorporate higher-torque brushless motors, a stiffer aluminium spine, onboard RGB-D cameras with deep convolutional neural networks \cite{li2020} for infrastructure-free navigation, and an upgraded controller using Model Predictive Control \cite{xin2020} or a compound disturbance observer \cite{tran2021} with tuned Kalman covariances \cite{murcia2016}.

\section*{Acknowledgment}
AI-assisted tools were used during the preparation of this manuscript to improve language, grammar, readability, and overall structure. We reviewed and revised all AI-assisted content and take full responsibility for the accuracy, originality, and final content of the manuscript.

\bibliographystyle{IEEEtran}
\bibliography{references}

\end{document}